\pdfoutput=1

\documentclass[letterpaper, 10pt, conference]{ieeeconf}   %

\IEEEoverridecommandlockouts                              %

\let\labelindent\relax                                    %

\usepackage{graphicx}               %
\usepackage{amsmath}                %
\usepackage{amssymb}                %
\usepackage{amsfonts}               %
\usepackage{enumitem}               %
\usepackage{siunitx}                %
\usepackage{url}                    %
\usepackage{xspace}                 %
\usepackage[T1]{fontenc}            %
\usepackage[hidelinks]{hyperref}    %
\usepackage{textcomp}               %
\usepackage{algorithm, algorithmic} %
\usepackage{array}                  %
\usepackage{eqparbox}               %
\usepackage{caption}                %

\newcommand{\BREAK}{\STATE \textbf{break}}

\makeatletter
\let\MYcaption\@makecaption
\makeatother
\usepackage[font=footnotesize]{subcaption}
\makeatletter
\let\@makecaption\MYcaption
\makeatother

\graphicspath{{images/}}
\DeclareGraphicsExtensions{.pdf,.jpeg,.png,.jpg}

\algsetup{linenosize=\small}

\providecommand{\abs}[1]{\lvert#1\rvert}
\providecommand{\norm}[1]{\lVert#1\rVert}

\newcommand{\R}{\mathbb{R}} %
\newcommand{\degreem}{^{\circ}} %

\newcommand{\seclabel}[1]{\label{sec:#1}}

\newcommand{\figlabel}[1]{\label{fig:#1}}
\newcommand{\alglabel}[1]{\label{alg:#1}}
\newcommand{\tablabel}[1]{\label{tab:#1}}
\newcommand{\eqnlabel}[1]{\label{eqn:#1}}

\newcommand{\figref}[1]{Fig.~\ref{fig:#1}\xspace}
\newcommand{\tabref}[1]{Table~\ref{tab:#1}\xspace}
\newcommand{\algref}[1]{Algorithm~\ref{alg:#1}\xspace}

\newcommand{\iguhop}{igus\textsuperscript{\tiny\circledR}$\!$ Humanoid Open Platform\xspace}

\newcommand{\opencv}{OpenCV\xspace}

\newcommand{\degree}{$\degreem$\xspace}

\title{\LARGE \bf A Monocular Vision System for Playing Soccer\\ in Low Color Information Environments}

\author{Hafez Farazi, Philipp Allgeuer, and Sven Behnke%
\thanks{All authors are with the Autonomous Intelligent Systems (AIS) Group, Computer Science Institute VI,
        University of Bonn, Germany. Email: {\tt\small farazi@ais.uni-bonn.de}. This work was partially funded
        by grant BE 2556/10 of the German Research Foundation (DFG).
        }}

\usepackage{eso-pic}

\AtBeginDocument{\AddToShipoutPictureFG*{\AtTextUpperLeft{\put(0,\LenToUnit{9pt}){\parbox{\textwidth}{\centering\bfseries
Proceedings of 10th Workshop on Humanoid Soccer Robots, International Conference on\\Humanoid Robots (Humanoids), Seoul, Korea, 2015
}}}}}
\begin{document}

\maketitle
\thispagestyle{empty}
\pagestyle{empty}

\begin{abstract}
Humanoid soccer robots perceive their environment exclusively through cameras.
This paper presents a monocular vision system that was originally developed for 
use in the RoboCup Humanoid League, but is expected to be 
transferable to other soccer leagues. Recent changes in the Humanoid League rules  
resulted in a soccer environment with less color coding than in 
previous years, which makes perception of the game situation more challenging. 
The proposed vision system addresses these challenges by using brightness and texture for the detection of the required field features and 
objects. Our system is robust to changes in lighting conditions, and is designed 
for real-time use on a humanoid soccer robot. This paper describes the main 
components of the detection algorithms in use, and presents experimental results 
from the soccer field, using ROS and the \iguhop as a testbed. The proposed 
vision system was used successfully at RoboCup 2015.
\end{abstract}

\section{Introduction}
\seclabel{introduction}

RoboCupSoccer is an ongoing effort to develop humanoid soccer robots, 
with the vision of them being able to win against the FIFA world champion team by 2050. Essential to 
this effort is the ability of the robots to perceive the game 
situation in real-time under realistic conditions. Each year, the RoboCup 
soccer leagues update their rules, to force participating teams 
to develop more advanced features for their robots, and to make the competitions 
more similar to real soccer games~\cite{Gerndt:RAM2015}. For the 2015 RoboCup, numerous changes were made in 
the Humanoid league rules that affect visual perception, including in particular the 
reduction of color coding on objects on the field. The ball is now only 
specified to be at least 50\% white, and the goal posts are now white instead of 
yellow. Moreover, the field lines---a major feature for localization on the field---are now painted onto artificial grass, and are as a result no longer white. Due to these changes, the simple color 
segmentation and blob detection approaches that were quite popular in the past 
\cite{laue2009efficient} \cite{farazi2014baset} have become unsuitable. This 
paper presents a monocular vision system that addresses the new challenges by relying more on 
relative brightness and texture. It was tested at RoboCup 2015 and found to perform well under the new rules.

\section{Related Work} 

Visual perception is a much researched topic within the RoboCup community, 
and many approaches have been developed in the past, albeit now for older 
versions of the rules. 

Strasdat et al.~\cite{Strasdat2006}, for example, developed a probabilistic method for robot localization relying on many unreliable detections of field features and field lines.
Schulz and Behnke~\cite{Schulz:AdvancedRobotics2012} demonstrated that the tracking of the 
robot pose is possible by using field lines only.

Many attempts have been made to reduce the reliance on color classification.
For example, Schulz et al.~\cite{Schulz2007} learned to detect the ball by classifying regions of interest with a neural network,
based on color contrast and brightness features.
Similarly, Metzler et al.~\cite{Metzler2011} learned the detection of Nao robots based on color histograms in regions of interest.

\begin{figure}[!tb]
\centering
\subcaptionbox{Convex hull of the green regions in the image (yellow), and the unwanted area (red arrows).\figlabel{field_detection_rawhull}}[0.48\linewidth][l]{\includegraphics[width=0.48\linewidth]{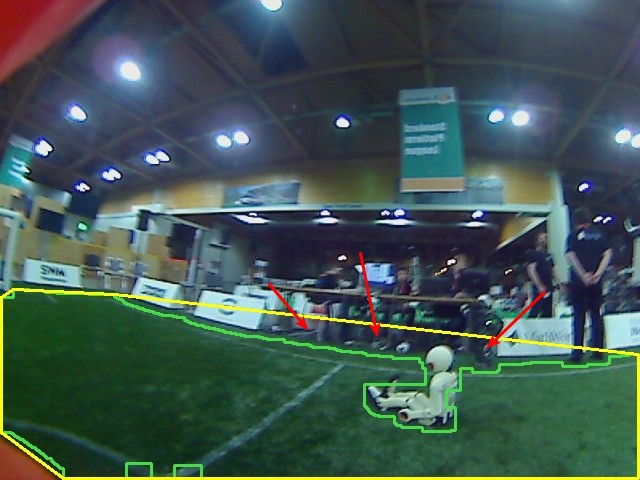}} \hspace{2pt}
\subcaptionbox{Boundary points of the undistorted convex hull. Green points are vertices of the raw extracted regions.\figlabel{field_detection_hullpts}}[0.48\linewidth][r]{\includegraphics[width=0.48\linewidth]{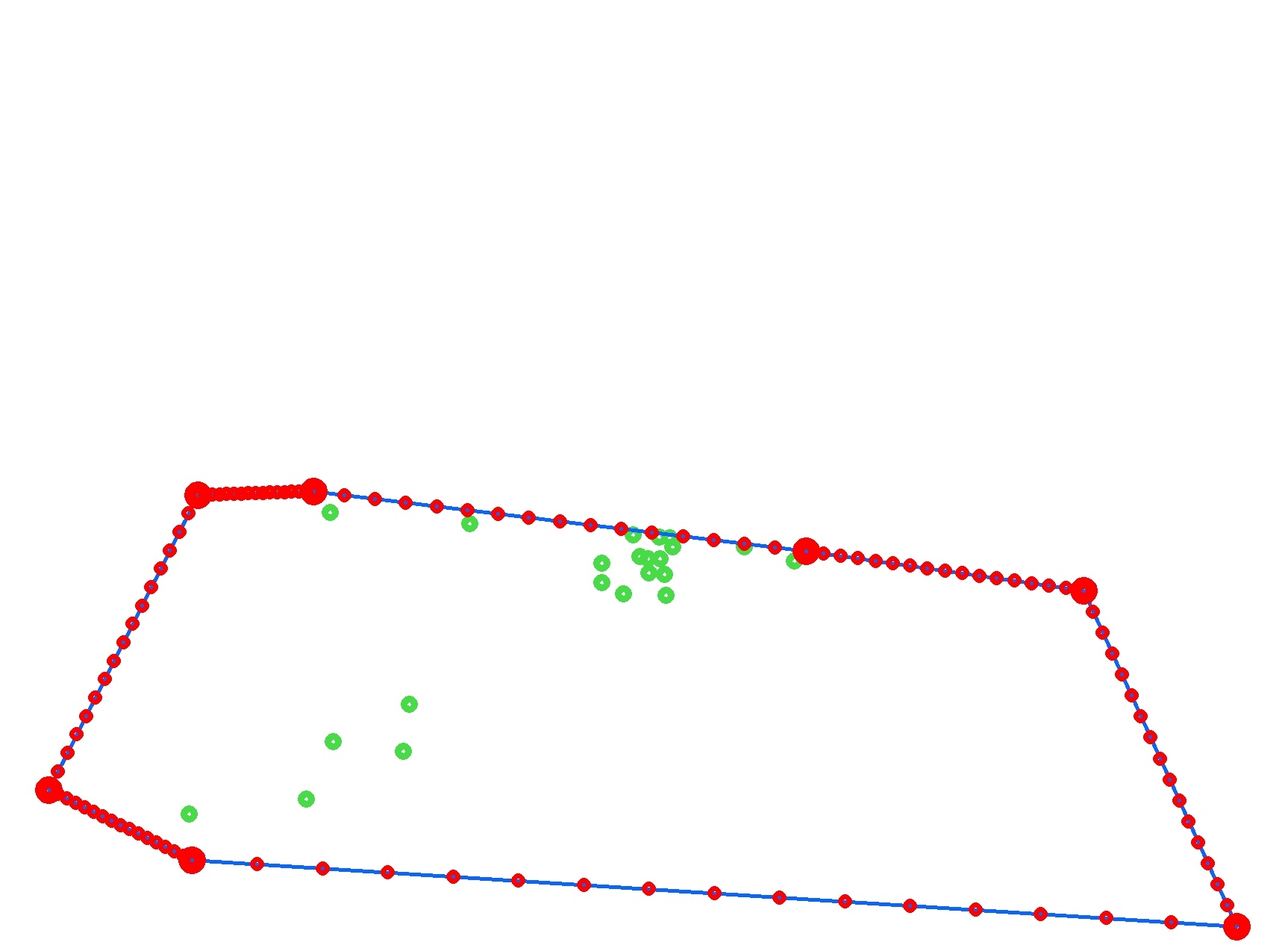}}
\hspace{3pt}
\subcaptionbox{Final detected field area.\figlabel{field_detection_final}}[0.48\linewidth][r]{\includegraphics[width=0.48\linewidth]{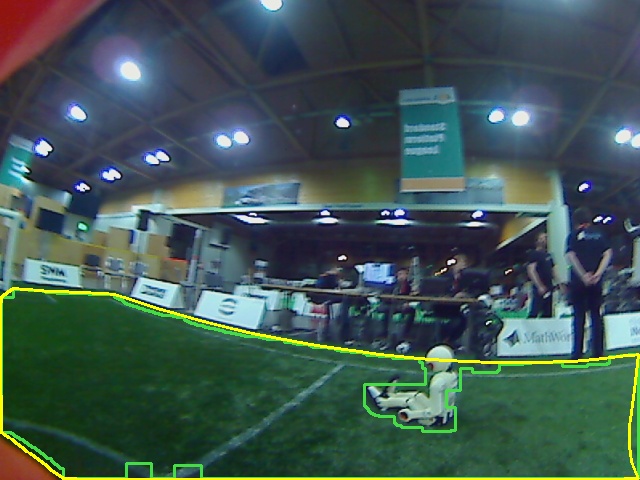}}
\subcaptionbox{An undistorted captured image.\figlabel{field_detection_undistort}}[0.48\linewidth][l]{\includegraphics[width=0.48\linewidth]{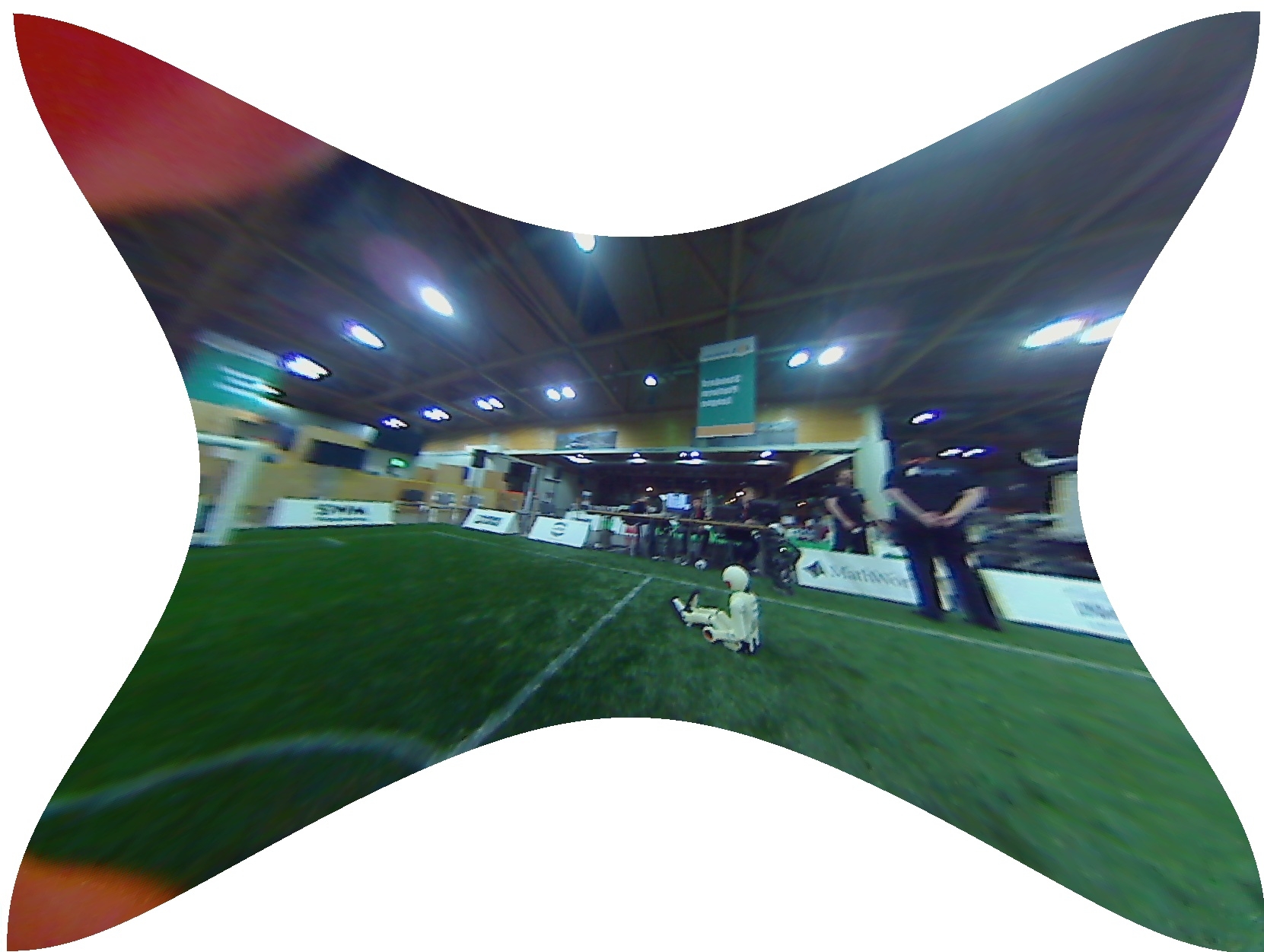}} \hspace{3pt}
\caption{Stages of field boundary detection.}
\figlabel{field_detection}
\end{figure}

More recent examples include the work of Houliston et al.~\cite{houlistonfast}, who 
introduced an adaptive lookup table in order to deal with changes in image 
illumination. Schwarz et al.~\cite{schwarzrobust} proposed a calibration-free 
vision system in the standard platform league (SPL), based on heuristics such as 
that the field is the maxima of the weighted histograms of each channel in the 
image. H{\"a}rtl et al.~\cite{hartl2014robust} presented work on color 
classification based on color similarity. To address the 
recent changes in goal color in the SPL, Cano et al.~\cite{robocupsympos2015} proposed a method for 
detecting white goals based on pixel intensities using the Y channel of the YUV 
color space. In order to cope with variations in illumination,
Reinhardt~\cite{reinhardt2011kalibrierungsfreie} proposed different heuristics that can be 
applied to color histograms. The middle size league (MSL) has used 
white goals for a number of years, but most approaches used in the MSL are not 
applicable to the humanoid league due to omnivision and active sensors such as 
Kinects being allowed.

The contributions of this paper include the introduction of several novel 
algorithms and techniques for performing soccer-related vision detections, and 
the development of a soccer vision system that can run in real-time and 
work under various lighting conditions, and in a low color information 
environment.

\section{System Overview}
\seclabel{system_overview}

In our vision system, the image is captured in the RGB color format, and 
converted to the HSV (Hue, Saturation and Value) color space using \opencv conversion routines~\cite{opencv_library}. For color 
classification, we use the HSV color space due to its intuitive nature, in particular its ability to separate brightness and 
chromatic information~\cite{caleiro2007color}. To ensure the short term 
consistency of the colors in the captured images, and to prevent unwanted abrupt 
image composition changes due to the camera itself, a fixed set of camera 
parameters, including for example brightness and exposure, are set in the camera 
device. In some cases, the camera may disconnect and reconnect from the USB bus, 
such as possibly when the robot falls. For this situation, the camera image 
acquisition thread is designed in such a way that it can automatically detect a 
disconnect event and reconnect to the camera device.

The vision process nominally acquires camera images at \SI{30}{\hertz}. Each 
captured image is labeled with a timestamp, along with the local robot position 
and orientation at that time. Feature detection routines are applied to each 
image, in the order \emph{field boundary}, \emph{ball}, \emph{lines}, 
\emph{circle}, \emph{goal posts}, and finally \emph{obstacle}. Due to the 
unwanted overhead of transferring images across process boundaries, all image 
acquisition and processing routines are implemented in a single ROS node, 
referred to as the vision module. The vision module publishes its detections and 
outputs to suitable ROS topics, which can then be used by other nodes.

\section{Camera Calibration}
\seclabel{camera_calib}

In the \iguhop, we use a wide-angle lens to allow more of the environment to be 
seen at once. This introduces significant distortion, however, which must be 
compensated when projecting image coordinates into egocentric world coordinates. 
The \opencv \cite{opencv_library} chessboard calibration procedure is used to 
determine intrinsic camera parameters that characterize the distortion. The 
following pinhole camera distortion model, including both radial and tangential 
distortion components, is used:
\begin{align}
a &= \frac{x}{z}, \quad b = \frac{y}{z}, \quad r^2 = a^2 + b^2, \eqnlabel{dist_abr}\\
\hat{x} &= a \Bigl(\! \tfrac{1 + k_1 r^2 + k_2 r^4 + k_3 r^6}{1 + k_4 r^2 + k_5 r^4 + k_6 r^6} \!\Bigr) + 2 p_1 a b + p_2 (r^2 + 2 a^2), \eqnlabel{dist_ab_to_xhat}\\
\hat{y} &= b \mspace{2mu} \Bigl(\! \tfrac{1 + k_1 r^2 + k_2 r^4 + k_3 r^6}{1 + k_4 r^2 + k_5 r^4 + k_6 r^6} \!\Bigr) + 2 p_2 a b + p_1 (r^2 + 2 b^2), \eqnlabel{dist_ab_to_yhat}\\
u &= c_x + f_x \hat{x}, \quad v = c_y + f_y \hat{y}, \eqnlabel{dist_uv}
\end{align}
where $(c_x, c_y)$ is the principal point, and $f_x$ and $f_y$ are the focal 
lengths, all in pixel units. $k_1$ to $k_6$ are the radial distortion 
coefficients, and $p_1$ and $p_2$ are the tangential distortion coefficients. 
The input $(x, y, z)$ is a world vector in the camera frame, and the output $(u, v)$
is the corresponding distorted image pixel position. For simplicity, 
higher-order coefficients and thin prism distortions are not included in the 
model. Due to limitations of the undistortion functionality in \opencv, 
producing poor results especially near the corners, a more accurate and 
efficient undistortion method was implemented based on the Newton-Raphson 
method. This method is used to populate a pair of lookup tables that allow 
$O(1)$ distortion and undistortion operations at runtime. The effect of 
undistortion is shown in \figref{field_detection_undistort}.

The distortion model allows projection to the camera frame, but further 
extrinsic parameters are required to allow projection to egocentric world 
coordinates. More specifically, at each point in time the transformation from 
the egocentric world frame to the camera frame must be known, which is both 
time- and robot-specific. We use the ROS-native tf2 library \cite{tf2} for this 
purpose, which amalgamates joint position and robot kinematics information to 
produce the required time-varying transforms. Although we have a relatively 
exact kinematic model of the robot, some variation still occurs in the real 
hardware, resulting in potentially large projection errors for distant objects. 
To account for this, we calibrate the kinematics on each robot using a 
hill-climbing approach that tunes translation and rotation offsets between the 
torso, which contains the IMU, and the camera. These offsets are crucial for 
good performance of the pixel-to-egocentric coordinate projection algorithm. The 
effect of the kinematic calibration is demonstrated in \figref{kin_calib}. For reference, the corresponding raw captured image is shown in 
\figref{line_detection_final}. Note that the calibration procedure is 
interactive in the sense that the user can select a number of points in the 
captured image and the corresponding true field locations. This is used as the 
input to the hill-climbing method, which seeks to minimize the reprojection 
error of the selected points. This completes the calibration of the camera for 
the purposes of the vision module.

\begin{figure}[b]
\centering
\subcaptionbox{Projected line and goal post detections prior to kinematic calibration, with projection errors marked.\figlabel{kin_calib_before}}[0.48\linewidth][l]{\includegraphics[width=0.48\linewidth]{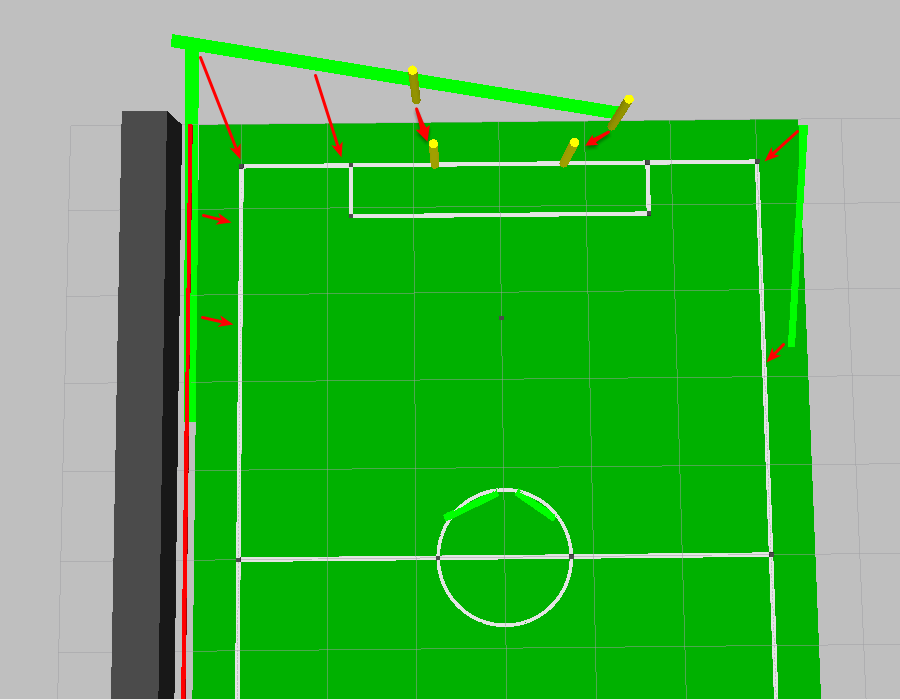}} \hspace{3pt}
\subcaptionbox{Projected line and goal post detections after kinematic calibration, with the field of view marked.\figlabel{kin_calib_after}}[0.48\linewidth][r]{\includegraphics[width=0.48\linewidth]{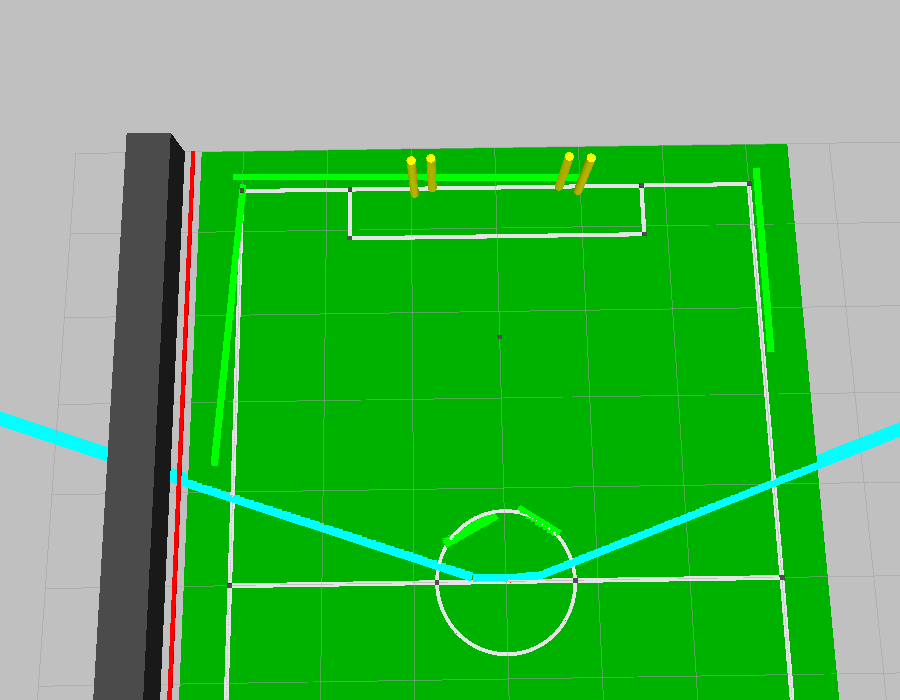}}
\caption{Effect of the kinematic calibration of the camera.}
\figlabel{kin_calib}
\end{figure}
\section{Feature Detection Algorithms}
\seclabel{feature_detection}

\subsection{Field Detection}
\seclabel{field_detection}

The soccer field used in RoboCup competitions is green, so field detection can 
be based on color segmentation in the HSV color space. Using a user-selected 
green color range, a binarized image is constructed, and all connected 
components that appear below the estimated horizon are found. Connected regions 
that have an area greater than some chosen threshold are taken into 
consideration for the extraction of a field boundary. If too many separate green 
regions exist, a set maximum number of regions are considered, prioritized by 
size and distance from the bottom of the image. Although it is a common approach 
to find a convex hull of all green areas directly in the image 
\cite{robocupsympos2015} \cite{laue2009efficient}, more care needs to be taken 
in our case due to the significant image distortion. 
\figref{field_detection_rawhull} shows how the convex hull may include parts of 
the image that are not the field. To exclude these unwanted areas, the vertices 
of the connected regions are first undistorted before calculating the convex 
hull, shown in \figref{field_detection_hullpts}. The convex hull points and 
intermediate points on each edge are then distorted back into the raw captured 
image, and the resulting polygon is taken as the field boundary. An example of 
the final detected field polygon is shown in \figref{field_detection_final}.

\subsection{Ball Detection}
\seclabel{ball_detection}

In previous years, most RoboCup teams used simple color segmentation and 
blob detection based approaches to find the orange ball. Now that the ball is 
mostly white, however, and generally with a pattern, such simple approaches no 
longer work effectively, especially since the lines and goal posts are also 
white. Our approach is divided into two stages. In the first stage, ball 
candidates are generated based on color segmentation, color histograms, shape, 
and size. White connected components in the image are found, and the 
Ramer-Douglas-Peucker algorithm \cite{ramer1972iterative} is applied to reduce 
the number of polygon vertices in the resulting regions. This is advantageous 
for quicker subsequent detection of circle shapes. The detected white regions 
are searched for at least one third full circle shapes within the expected 
radius ranges using a technique very similar to \algref{circle_detection}. Color 
histograms of the detected circles are calculated for each of the three HSV 
channels, and compared to expected ball color histograms using the Bhattacharyya 
distance \cite{opencv_library}. Circles with a suitably similar color 
distribution to expected are considered to be ball candidates.

\begin{figure}[b]
\parbox{\linewidth}{\centering
\includegraphics[width=0.088\linewidth]{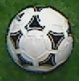}
\includegraphics[width=0.088\linewidth]{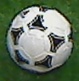}
\includegraphics[width=0.088\linewidth]{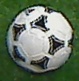}
\includegraphics[width=0.088\linewidth]{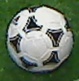}
\includegraphics[width=0.088\linewidth]{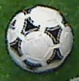}
\includegraphics[width=0.088\linewidth]{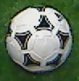}
\includegraphics[width=0.088\linewidth]{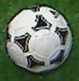}
\includegraphics[width=0.088\linewidth]{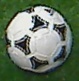}
\includegraphics[width=0.088\linewidth]{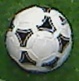}
\includegraphics[width=0.088\linewidth]{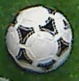}}
\caption{Extending one positive sample (leftmost) to ten, by applying rotations and mirroring operations.}
\figlabel{ball_samples}
\end{figure}
\begin{figure}[!tb]
\centering
\subcaptionbox{Ball detection in an undistorted image, with other white objects.\figlabel{ball_detection_lab}}[0.5025\linewidth][l]{\includegraphics[width=0.5025\linewidth]{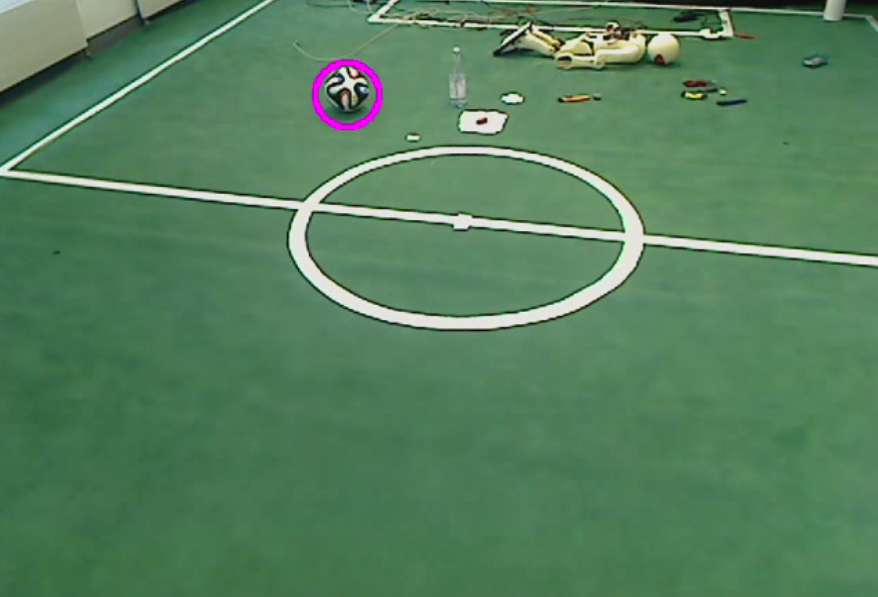}} \hspace{3pt}
\subcaptionbox{Multiple balls detected on a soccer field, with image distortion.\figlabel{ball_detection_field}}[0.455\linewidth][r]{\includegraphics[width=0.455\linewidth]{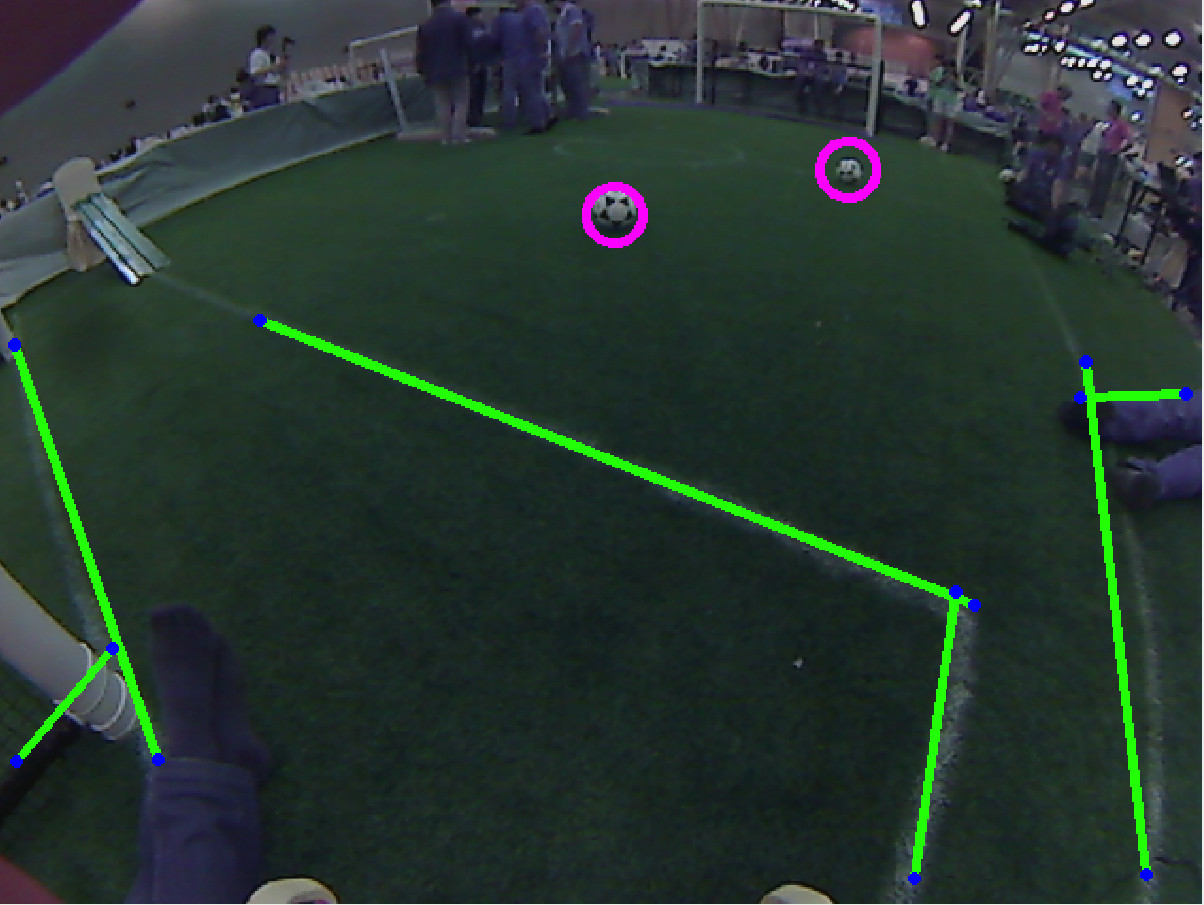}}
\\
\caption{Ball detections under various conditions.}
\figlabel{ball_detection}
\end{figure}
\begin{algorithm}[!tb]
\caption{Merge similar line segments.}
\alglabel{merge_all_line_segments}
\begin{algorithmic}[1]
\REQUIRE A set of line segments $S$
\ENSURE  A set of line segments $N$ where $\abs{N} \leq \abs{S}$ 
  \STATE $N \leftarrow S$
  \REPEAT
    \STATE $M \leftarrow N$
    \STATE $m \leftarrow \abs{M}$
    \STATE $N \leftarrow \emptyset$
    \WHILE{$M \neq \emptyset$}
      \STATE $X \leftarrow$ Any element of $M$
      \STATE $M \leftarrow M \setminus \{X\}$
      \STATE $\hat{m} \leftarrow \abs{M}$
      \FOR{$Y \in M$}
        \IF {$MinDistance(X,Y) < \SI{0.25}{\metre}$ \\ \AND $AngleDif\!f(X,Y) < 15$\degree}
          \STATE $N \leftarrow N \cup \{Merge(X,Y)\}$
          \STATE $M \leftarrow M \setminus \{Y\}$
          \BREAK
        \ENDIF
      \ENDFOR
      \IF{$\abs{M} = \hat{m}$}
        \STATE $N \leftarrow N \cup \{X\}$
      \ENDIF
    \ENDWHILE
  \UNTIL{$\abs{N} = m$}
  \RETURN $N$
\end{algorithmic}
\end{algorithm}
\begin{algorithm}[!tb]
\caption{$Merge(X,Y)$: Merge two line segments.}
\alglabel{merge_two_line_segments}
\begin{algorithmic}[1]
\REQUIRE Two line segments $X$ and $Y$
\ENSURE  Merged line segment $Q$
  \STATE $\theta \leftarrow 0$
  \STATE $X_m = Midpoint(X)$
  \STATE $Y_m = Midpoint(Y)$
  \STATE $r \leftarrow \norm{X} / (\norm{X} + \norm{Y})$
  \STATE $P \leftarrow r X_m + (1-r) Y_m$
  \IF{$\norm{X} \geq \norm{Y}$}
    \STATE $\theta \leftarrow Slope(X)$
  \ELSE
    \STATE $\theta \leftarrow Slope(Y)$
  \ENDIF
  \STATE $Z \leftarrow$ The line through point $P$ of slope $\theta$
  \STATE $S \leftarrow$ Orthogonal projections of the endpoints of $X$ and \\$\mspace{36mu}$ $Y$ onto $Z$
  \STATE $Q \leftarrow$ The largest line segment defined by points in $S$
  \RETURN $Q$
\end{algorithmic}
\end{algorithm}

In the second stage of processing, a dense histogram of oriented gradients (HOG) 
descriptor \cite{dalal2006object} is applied in the form of a cascade 
classifier, with use of the AdaBoost technique. Using this cascade classifier, 
we reject those candidates that do not have the required set of HOG features. 
The AdaBoost technique is used because it provides a powerful bound on the 
generalization performance \cite{schapire1998boosting}. In contrast to what is 
suggested in \cite{dalal2006object} however, which was targeted at pedestrian 
detection, we do not use a multi-scale sliding window technique. Instead, to 
save computational time, we only apply the HOG descriptor to the regions 
suggested by the ball candidates. The aim of using the HOG descriptor is to find 
a description of the ball that is largely invariant to changes in illumination 
and lighting conditions. In contrast to other common feature descriptors, such 
as SIFT \cite{sift}, HOG features are easy to visualize 
\cite{vondrick2013hoggles}, and have for example been found to be more reliable 
than SIFT in pedestrian detection \cite{dalal2006object}. The HOG descriptor is 
not rotation invariant however, so to detect the ball from all angles, and to 
minimize the user's effort in collecting training examples, each positive image is 
rotated by \textpm{}10\degree and \textpm{}20\degree, selectively mirrored, with 
the resulting images being presented as new positive samples. Greater rotations 
are not considered to allow the cascade classifier to learn the shadow under the 
ball. So in summary, each positive sample that the user provides is extended to 
a total of 10 positive samples, as shown in \figref{ball_samples}. With a set of about 400 positive and 
700 negative samples that are gathered directly from the robot camera and with 
20 weak classifiers, training takes approximately 10 hours.

As demonstrated in \figref{ball_detection}, the described approach can detect 
balls with very few false positives, even in environments cluttered with white 
and with varying lighting conditions. In our experiments, we found that this 
approach can detect the FIFA size 4 ball up to four meters away. An interesting 
result is that our approach can find the ball in undistorted and distorted 
images with the same classifier. A full set of generalized Haar wavelet features 
\cite{haar} was tested on the same data set, and although the result was 
comparable, the training time was about 10 times slower than for the HOG 
descriptor. A Local Binary Patterns (LBP) feature classifier 
\cite{liao2007learning} was implemented in another comparative test, but the 
ball detection results were poor.

\begin{table}[!tb]
\renewcommand{\arraystretch}{1.3}
\caption{Helper functions used in the algorithms.}
\tablabel{function_list}
\centering
\begin{tabular}{r | p{0.52\linewidth}}
\hline
$MinDistance(X,Y)$ & Returns the minimum distance between two line segments $X$ and $Y$\\
$AngleDif\!f(X,Y)$ & Returns the angle between two line segments in degrees\\
$Merge(X,Y)$ & See \algref{merge_two_line_segments}\\
$Midpoint(X)$ & Returns the midpoint of a line segment\\
$Slope(X)$ & Returns the slope of a line segment\\
$MaxPtDistance(S)$ & Returns the maximum distance between any pair of points in the set $S$\\
$GetBisector(X)$ & Returns the perpendicular bisector line of the given line segment\\
$HasIntersection(X,Y)$ & Returns whether two line segments intersect at a point\\ 
$GetIntersection(X,Y)$ & Returns the intersection point of two line segments\\
$DistanceToLine(P,X)$ & Returns the Euclidean distance between point $P$ and line segment $X$\\
$Mean(S)$ & Returns the mean of the points in the set $S$\\
\hline
\end{tabular}
\end{table}
\subsection{Field Line Detection}
\seclabel{line_detection}

Due to the introduction of artificial grass in the RoboCup humanoid league, the 
lines are no longer clearly visible and identifiable on the field. In past 
years, many teams based their line detection approaches on the segmentation of 
the color white. This is no longer a robust approach due to the increased number 
of white objects on the field, and due to the visual variability of the lines. 
Our approach is to detect spatial changes in brightness in the image using a 
canny edge detector \cite{canny1986computational} on the V channel of the HSV 
color space. The V channel encodes brightness information, and the result of the 
canny edge detector on this input is quite robust to changes in lighting 
conditions, so there is no need to tune parameters for changing lighting 
conditions. A probabilistic Hough line detector \cite{matas2000robust} is used 
to extract line segments of a certain minimum size from the detected edges. This 
helps to reject edges from white objects in the image that are not lines. The 
output line segments are filtered in the next stage to avoid false positive line 
detections where possible. A thresholding technique is used to try to ensure 
that the detected lines cover white pixels in the image, have green pixels on 
either side, and are close on both sides to edges returned by the edge detector. 
The last of these checks is motivated by the expectation that white lines, in an 
ideal scenario, will produce a pair of high responses in the edge detector, one 
on each border of the line. Ten equally spaced points are chosen on each line 
segment under review, and two normals to the line are constructed at each of 
these points, of approximate \SI{5}{\centi\metre} length in each of the two 
directions. The pixels in the captured image underneath these normals are 
checked for white color and green color, and the output of the canny edge 
detector is checked for high response. The number of instances where these three 
checks succeed are independently totaled, and if all three counts---$V_W$, 
$V_G$ and $V_E$ respectively---exceed the configured thresholds, the line segment 
is accepted, otherwise the line segment is rejected.

In the final stage, similar line segments are merged with each other, as 
appropriate, to produce fewer and bigger lines, as well as cover those line 
segments that might be partially occluded by another robot. Prior to merging, we 
project each line segment to the egocentric world coordinates. The merging 
algorithm is detailed in \algref{merge_all_line_segments}, which in turn uses 
\algref{merge_two_line_segments} to perform the individual merge operations. A 
list of the simple helper functions that are used in the algorithms in this 
paper is presented in \tabref{function_list}. Note that the constant thresholds 
used in \algref{merge_all_line_segments}, are indicating the maximum allowed 
difference in angle and distance between two lines to be merged, and the numeric 
values are suitable for the current humanoid league field dimensions.

The final result is a set of line segments that relates to the lines and center 
circle on the field. Line segments that are under a certain threshold in length 
undergo a simple circle detection routine, detailed in 
\algref{circle_detection}, to find the location of the center circle. In our 
experiments, we found that this approach can detect circle and line segments up 
to \SI{4.5}{\metre} away.

\begin{figure}[!tb]
\centering
\subcaptionbox{The output of the probabilistic Hough line detection on the extracted edges (thin red lines).\figlabel{line_detection_raw}}[0.48\linewidth][l]{\includegraphics[width=0.48\linewidth]{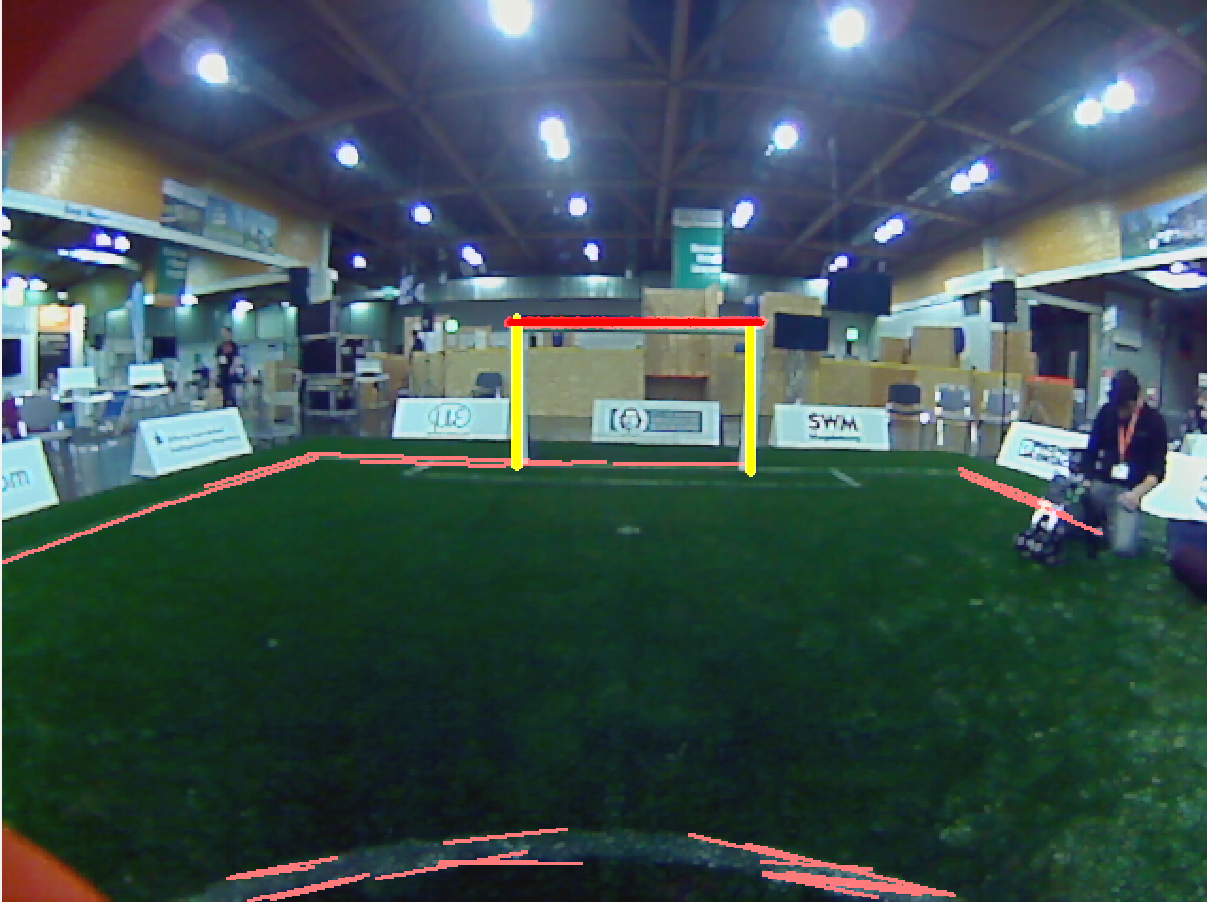}} \hspace{3pt}
\subcaptionbox{The final field line (green lines) and goal post (yellow lines) detections, marked in the captured image.\figlabel{line_detection_final}}[0.48\linewidth][r]{\includegraphics[width=0.48\linewidth]{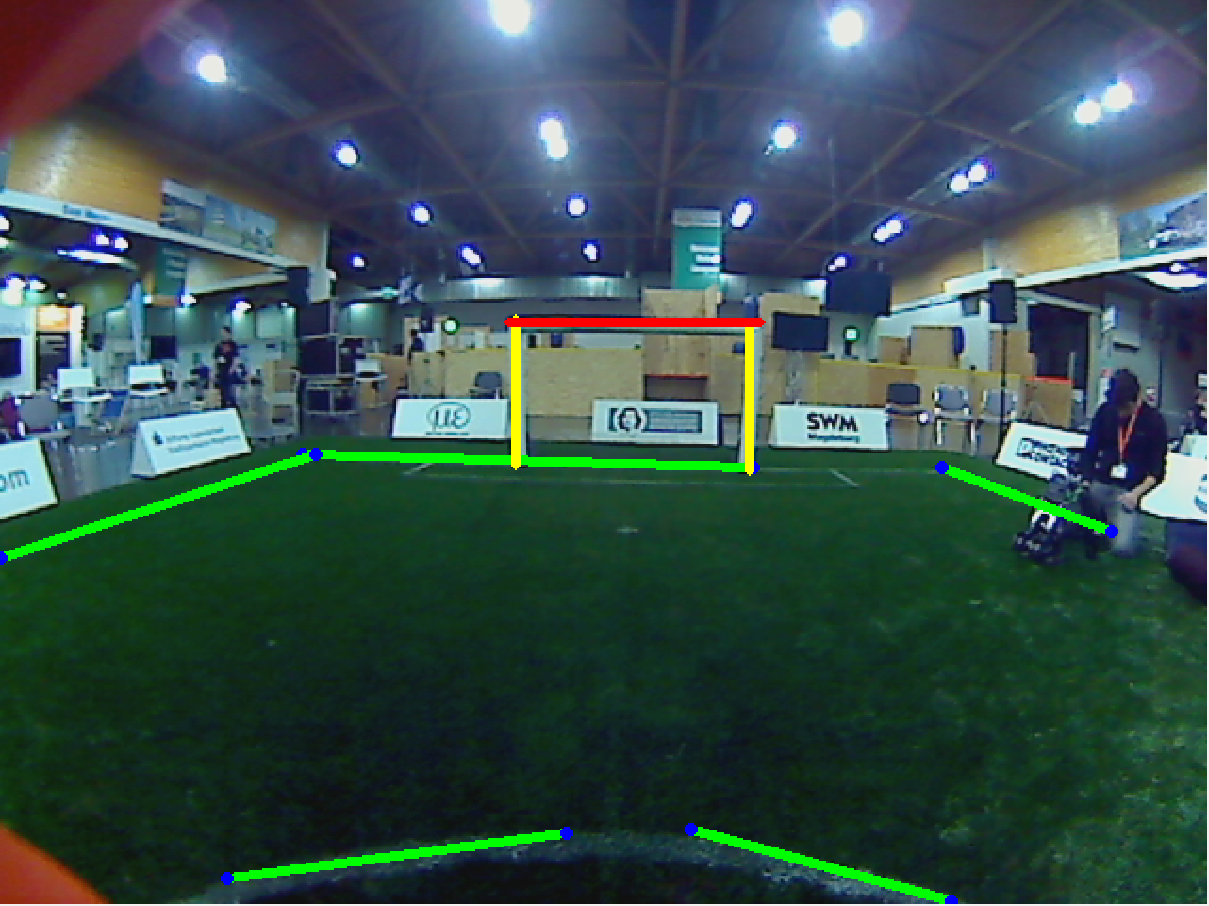}}
\\
\caption{Line detections on a soccer field.}
\figlabel{line_detection}
\end{figure}

We generate a probability estimation for each line detection, so that a 
localization algorithm for example can assign higher weights to detections of 
higher probability. For a detected line $X$ of length $L$, and check counts 
$V_G$ and $V_E$, the probability $P(X)$ of the detection is calculated as
\begin{equation}
P(L, V_G, V_E) = 
\begin{cases}
\frac{V_G}{V_{GM}} \cdot \frac{V_E}{V_{EM}} \cdot \frac{L-1}{3} & \text{if $1 < L < 3 $}, \\[2pt]
\frac{V_G}{V_{GM}} & \text{if $L \geq 3$}, \\[2pt]
0 & \text{if $L \leq 1 $},
\end{cases}
\eqnlabel{line_prob}
\end{equation}
where $V_{GM}$ is the maximum possible value of $V_G$, and $V_{EM}$ is the 
maximum possible value of $V_E$. For a detected circle $C$, consisting of line 
segments $X_i$ for $i = 1\ldots n$, with corresponding check counts $V_{Gi}$ and 
$V_{Ei}$, the probability $P(C)$ of the detection is calculated as
\begin{equation}
P(C) = \frac{\:\sum\limits_{i=1}^n {V_G}_i\:}{n V_{GM}} \cdot \frac{\:\sum\limits_{i=1}^n {V_E}_i\:}{n V_{EM}} \cdot \frac{MaxPtDistance(P)}{0.75},
\eqnlabel{circle_prob}
\end{equation}
where $P$ is the set of estimated circle centre points, as calculated by 
\algref{circle_detection}.

\begin{algorithm}[!tb]
\caption{Detect circle from line segments.}
\alglabel{circle_detection}
\begin{algorithmic}[1]
\REQUIRE A set of line segments $S$
\ENSURE  A boolean flag whether a circle is detected  
\ENSURE  The centre $C \in \R^2$ of the detected circle  
  \STATE $P \leftarrow \emptyset$
  \FORALL{Line segments $X,Y \in S$ such that $X \neq Y$}
    \STATE $X_B \leftarrow$ $GetBisector(X)$
    \STATE $Y_B \leftarrow$ $GetBisector(Y)$
    \IF{$HasIntersection(X_B,Y_B)$}
      \STATE $C \leftarrow GetIntersection(X_B,Y_B)$
      \IF{$DistanceToLine(C,X) \approx 0.75$ \\ \AND $DistanceToLine(C,Y) \approx 0.75$} 
        \STATE $P \leftarrow P \cup \{C\}$
      \ENDIF
    \ENDIF
  \ENDFOR
  \IF {$\abs{P} \geq 5$ \AND $MaxPtDistance(P) < 0.75$} 
    \RETURN $\TRUE, \,Mean(P)$
  \ENDIF
  \RETURN $\FALSE, \,(0,0)$
\end{algorithmic}
*: Note that the number $0.75$ in this algorithm is related to the expected circle radius in metres.
\end{algorithm}
\subsection{Goal Detection}
\seclabel{goal_detection}

In our vision module, detection of the white goal posts is divided into two 
stages. In the first stage, the image is binarized using color segmentation of 
the color white, defined as a particular region of the HSV color space by the 
user. A sample output is shown in \figref{goal_detection_binary}. Horizontal and 
vertical lines are then extracted from the binarized image using probabilistic 
Hough line detection \cite{matas2000robust}. Using a similar approach as for 
field line detection, the detected vertical and horizontal line segments are 
merged together to produce fewer, bigger lines. The main idea behind merging is 
to have one line segment on each goal post, instead of two on each side. 
Moreover, by merging line segments we can overcome only partially observable 
goal posts, and motion-blurred images in which the background may blend into 
some part of the goal post. The result of the merging is shown in 
\figref{goal_detection_merged}. In the second stage of goal post detection, each 
of the vertical line segments that do not meet the following criteria are 
rejected.
\begin{itemize}
 \item The length of the line segment must be within a certain range, dependent 
       on the distance from the robot to the projected bottom point of the line.
 \item The bottom of the line must be within the field.
 \item The top of the line must be above the estimated horizon.
 \item On a kid-size field, if the goal post candidate is more than 
       \SI{2}{\metre} from the robot there must be one horizontal line segment
       close to the candidate.
\end{itemize}
To reject the vertical line segments that might belong to a standing white 
robot, we check the homogeneity of the texture of each of the merged vertical 
candidates. To do this, we calculate 10 equally spaced points on the line, and 
search locally around each of these points for large changes in the H channel of 
the HSV color space. If many changes in the H channel are observed, the 
candidate is rejected. Experiments have shown that this approach can detect goal 
posts up to \SI{5}{\metre} away.

As for field lines, we assign a probability $P(G)$ to detected goal posts, using 
the equation
\begin{equation}
P(G) = 
\begin{cases}
\tfrac{3}{4} & \text{if $n = 1$}, \\
1 - \tfrac{1}{6} \bigl\lvert D - \norm{X_1 - X_2} \bigr\rvert & \text{if $n = 2$}, \\
1 - \frac{n}{5} & \text{if $3 \leq n \leq 4$}, \\
0 & \text{if $n \geq 5$}, \\
\end{cases}
\eqnlabel{goal_prob}
\end{equation}
where $n$ is the number of the detected goal posts, $D$ is the known goal width, 
and $X_1, \ldots, X_n$ are the egocentric world vectors to the $n$ detected goal 
posts.

\begin{figure}[!tb]
\centering
\subcaptionbox{White color segmented image in which pixels that are not considered to be white have been painted black.\figlabel{goal_detection_binary}}[0.48\linewidth][l]{\includegraphics[width=0.48\linewidth]{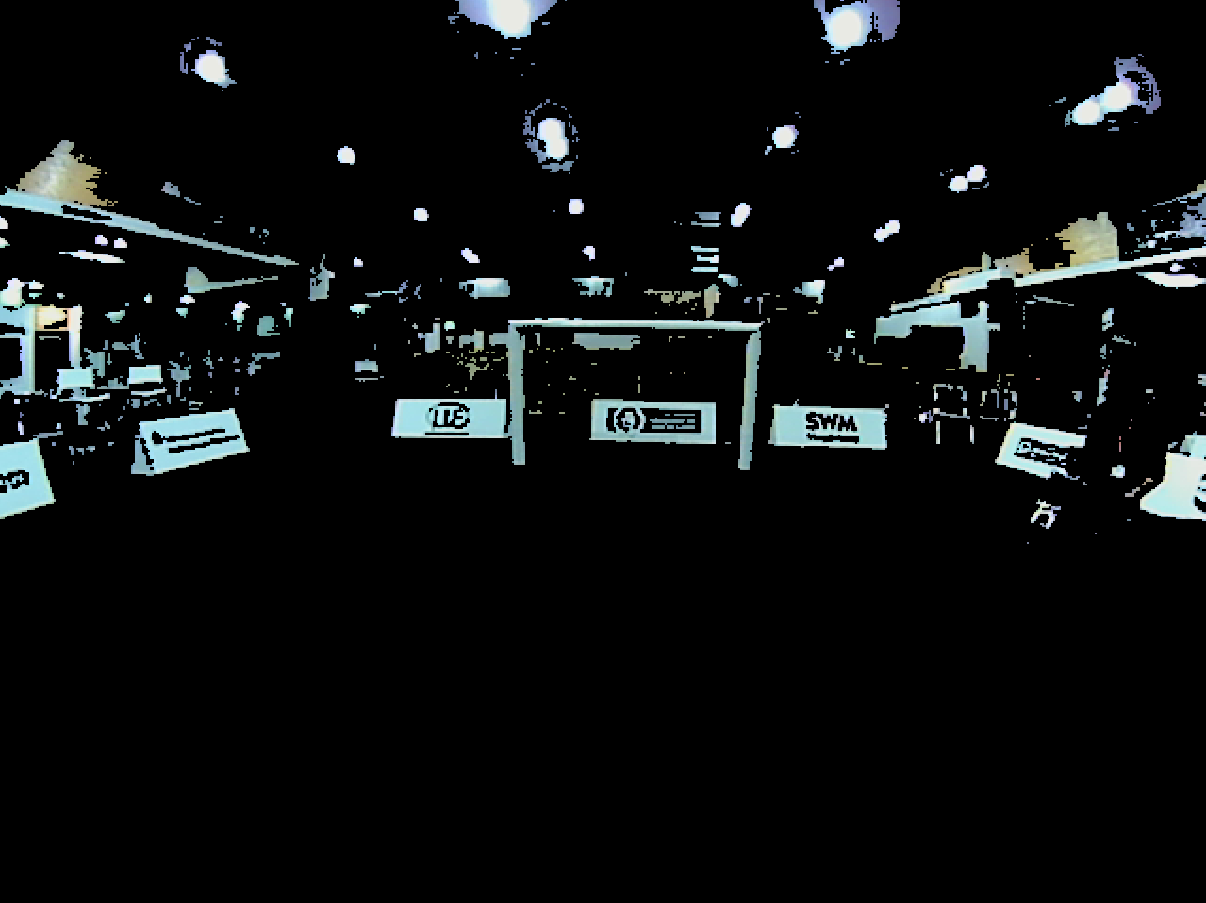}} \hspace{3pt}
\subcaptionbox{Result after merging all extracted horizontal and vertical white line segments (annotated in blue).\figlabel{goal_detection_merged}}[0.48\linewidth][r]{\includegraphics[width=0.48\linewidth]{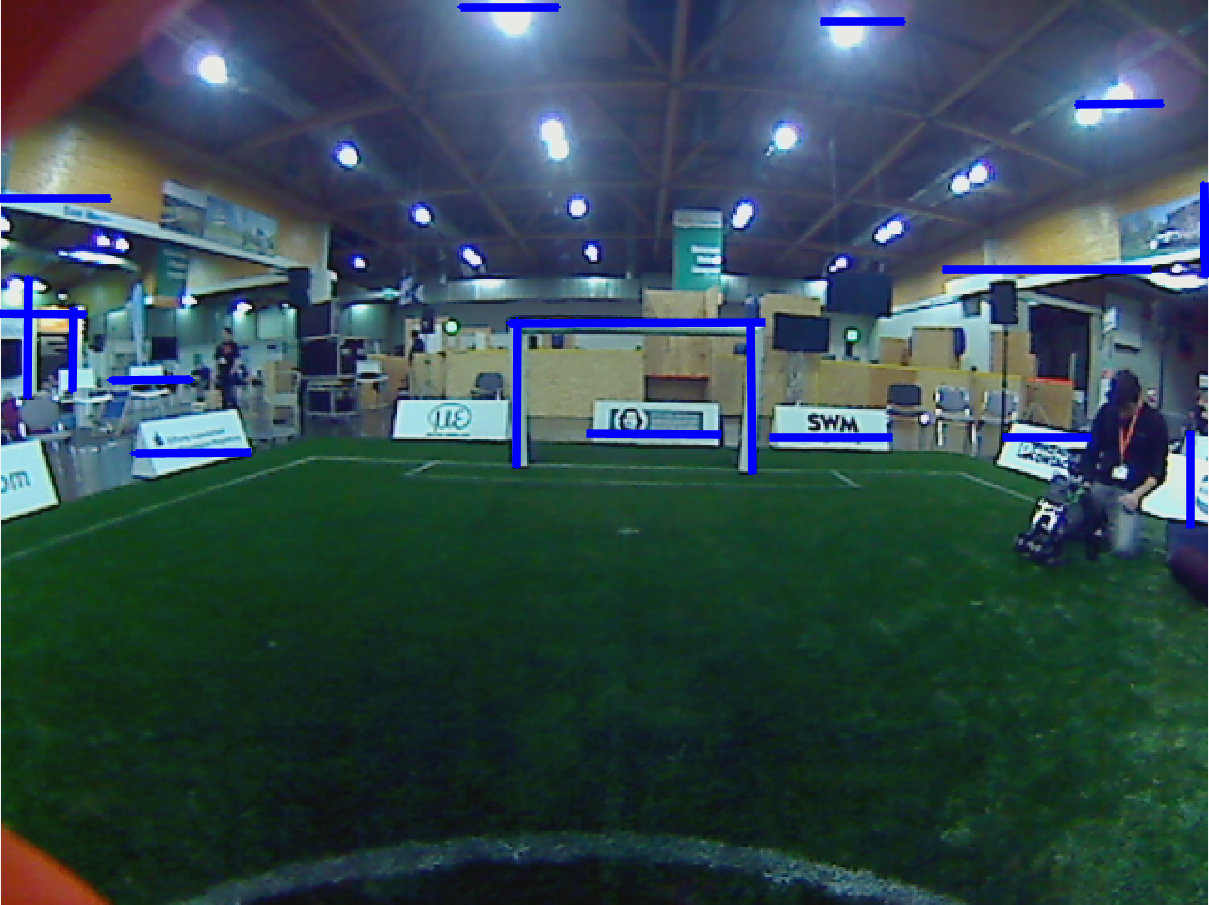}}
\\
\caption{Goal detection on a soccer field.}
\figlabel{goal_detection}
\end{figure}
\subsection{Obstacle Detection}
\seclabel{obstacle_detection}

The rules of the RoboCup humanoid league state that all robots should have 
mostly black feet. As such, obstacle detection is based on color segmentation of 
the black color range that is defined by the user. We search for black connected 
components within the field boundary, and if a detected component is large 
enough and within a certain predefined distance interval from the robot, the 
lowest point of the component is returned as the egocentric world coordinates of 
an obstacle. To prevent the detection of own body parts as obstacles, an image 
mask is implemented that is dependent on the position of the head. This allows 
the regions of the image that are expected to contain parts of the own body to 
be ignored. Using this approach, we are able to detect robots at distances of up 
to \SI{2}{\metre}.

\section{Experimental Results}
\seclabel{experimental_results}

We used the \iguhop \cite{Allgeuer2015} in our experiments. This robot is about 
\SI{90}{\centi\metre} tall and can be used in both the teen-size and kid-size 
leagues. The robot has a computer with a dual-core AMD E-450 \SI{1.65}{GHz} 
processor and 2GB of memory. This robot is equipped with two 720p Logitech C905 
USB cameras. Each camera is fitted with a wide-angle lens that has an infrared 
cut-off filter, yielding a diagonal field of view of approximately 150\degree. 
We use the Video4Linux2 library to capture images in RGB format at a resolution 
of $640 \times 480$. So far, we have only evaluated the vision system on one 
camera at a time. During RoboCup 2015, the computation time of the vision system 
was measured on the robot. Using only one thread, an average cycle time of 
\SI{25}{\milli\second} was achieved, with a minimum of \SI{15}{\milli\second}, 
and maximum of \SI{38}{\milli\second}, so the proposed vision system can easily 
run on a single CPU core at \SI{30}{\hertz}. Of all the components of the vision 
system, the ball detection is the most time consuming, and the robot detection 
is the least.

We assessed the performance of the proposed vision system both qualitatively and 
quantitatively, using an approach similar to that used by Schwarz et al.~\cite{schwarzrobust}. As inputs, we used recorded data from the RoboCup German 
Open 2015, which included captured image data from various different field 
locations and lighting conditions. The recorded data contained image data, 
kinematic information and dead-reckoning walking data. Many captured images were 
affected by motion blur due to walking and head panning motions. The data was 
evaluated frame by frame by the authors to assess the performance of the vision 
system in real soccer situations. \tabref{criteria} defines a set of detection 
criteria for the evaluation of the recorded bags. The success rate of the 
detections was calculated given the criteria, and the number of false positives 
was recorded. The results are shown in \tabref{results}. Despite the fact that 
Schwarz et al. tested their vision algorithms in a more color-coded SPL 
environment \cite{schwarzrobust}, we achieve comparable results, and in fact 
achieve better results when it comes to circle, ball and robot detections. In 
another comparison of the obtained results, we found that our method outperforms 
that of H{\"a}rtl et al. \cite{hartl2014robust} in terms of both line and circle 
detections, even though their results were obtained in the previous year's more 
color-coded RoboCup environment. 

In a second experiment, we tested our vision 
system under various different lighting conditions, including normal, very poor, 
ambient and natural lighting conditions, without tuning any parameters. The 
results are shown in \figref{light}. Consistent detections can be seen in all 
scenarios. We claim that the achieved result is sufficient for successful soccer 
game play in the new low color information RoboCup environment.

\begin{figure}[!tb]
\centering
\subcaptionbox{Normal lighting conditions.\figlabel{light5}}[0.48\linewidth][l]{\includegraphics[width=0.48\linewidth]{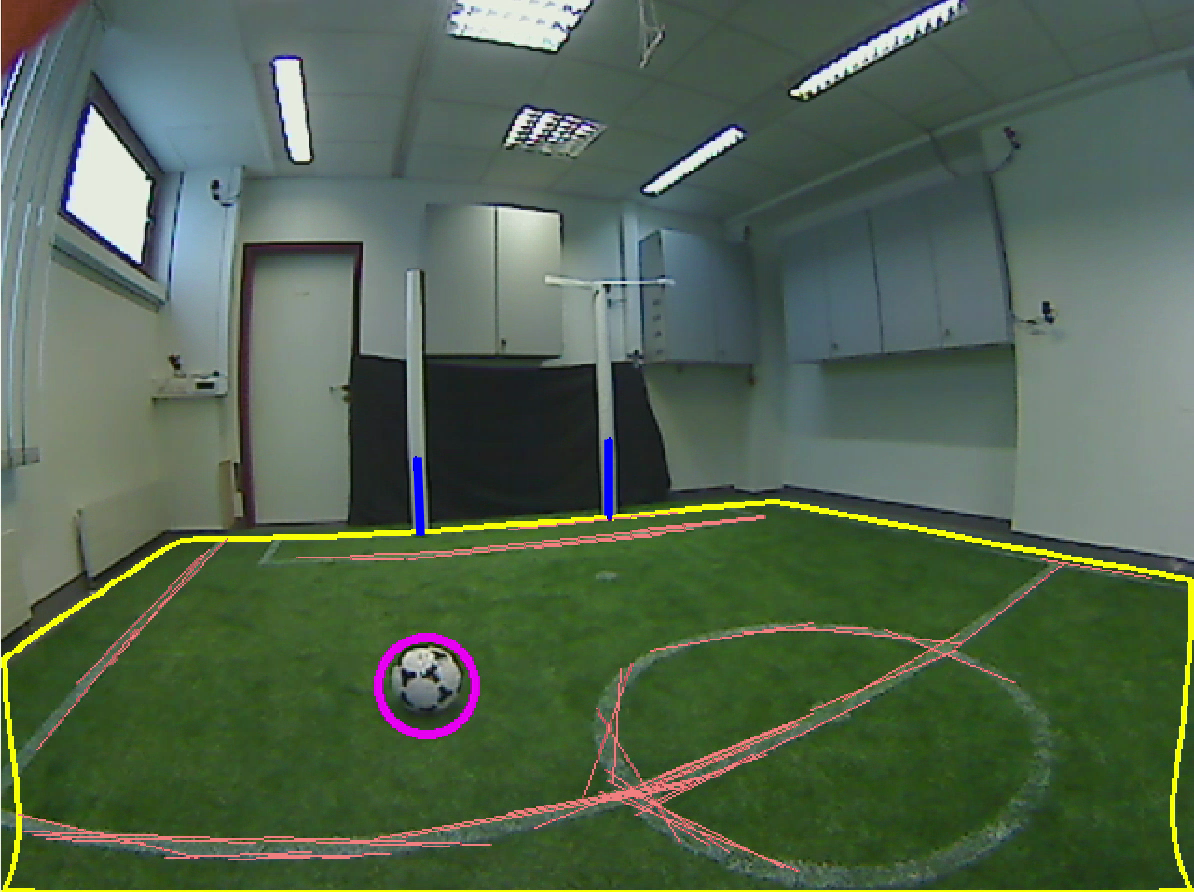}} \hspace{2pt}
\subcaptionbox{Poor lighting conditions.\figlabel{light3}}[0.48\linewidth][r]{\includegraphics[width=0.48\linewidth]{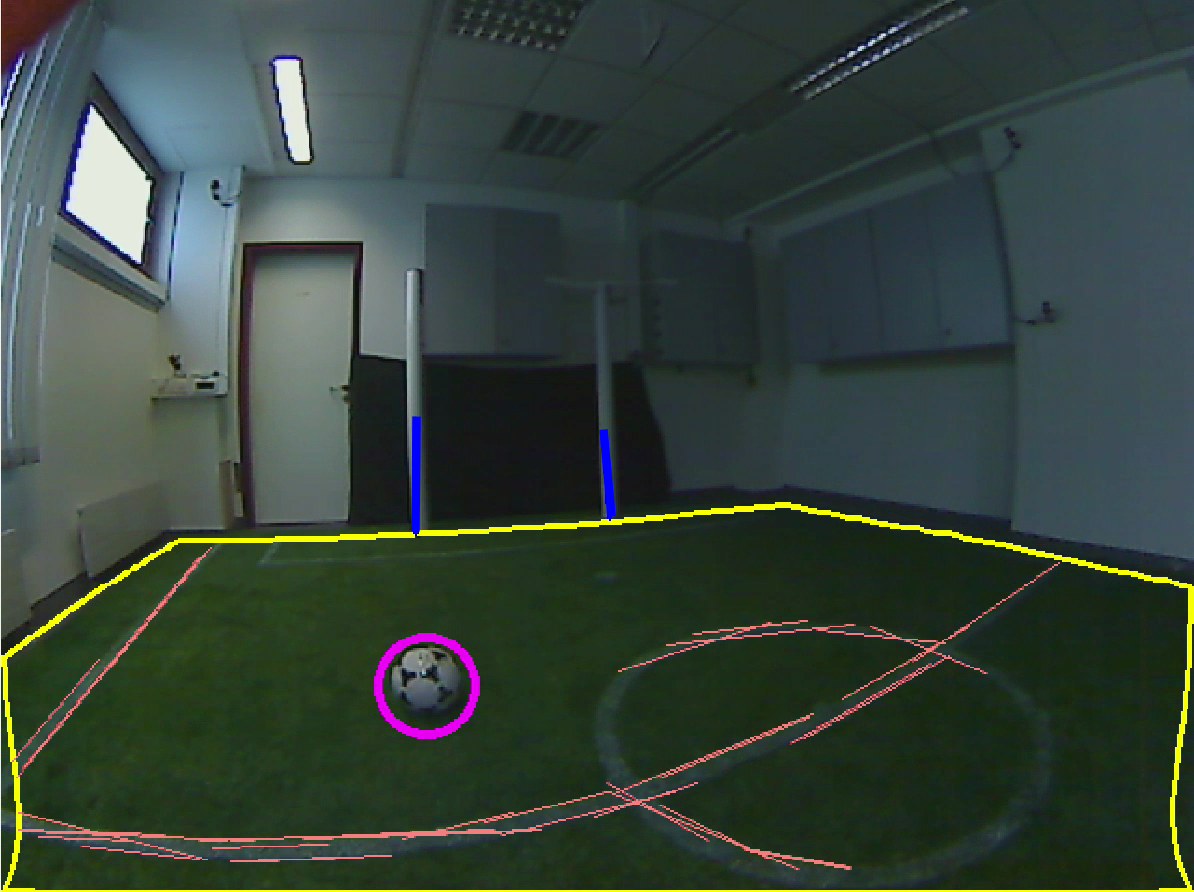}}
\\\medskip
\subcaptionbox{Natural lighting.\figlabel{light2}}[0.48\linewidth][l]{\includegraphics[width=0.48\linewidth]{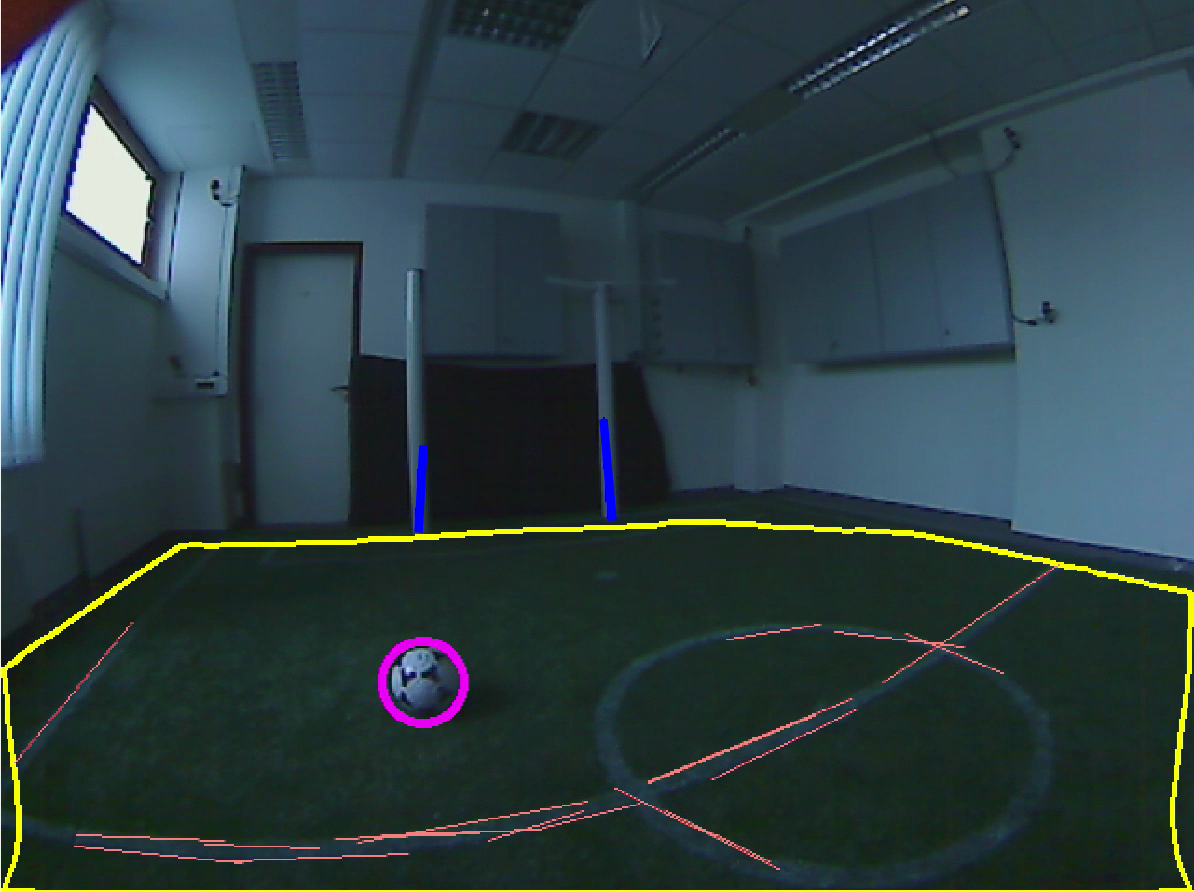}} \hspace{2pt}
\subcaptionbox{No ambient lighting.\figlabel{light1}}[0.48\linewidth][r]{\includegraphics[width=0.48\linewidth]{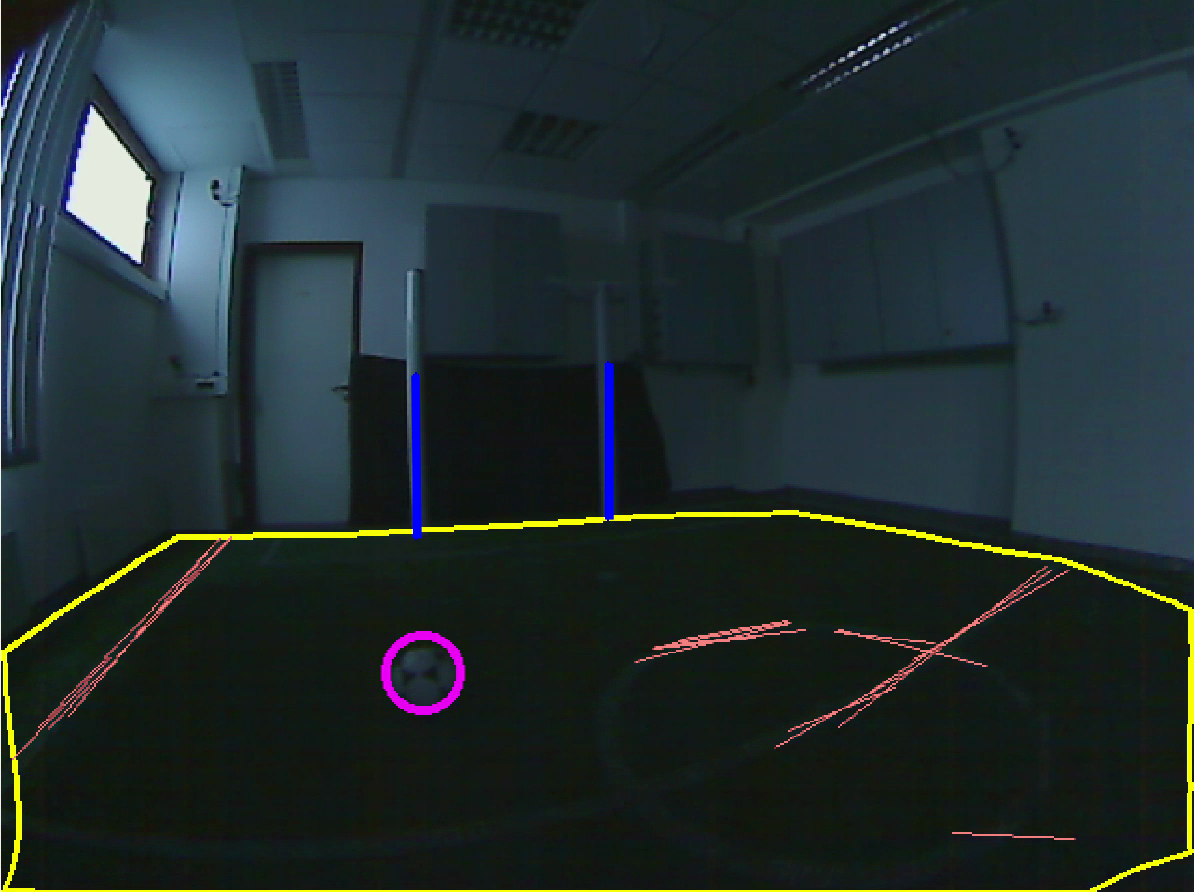}}
\\
\caption{Experimental results under various lighting conditions without any changes in the algorithm parameters.}
\figlabel{light}
\end{figure}
\begin{table}[!tb]
\renewcommand{\arraystretch}{1.3}
\caption{Evaluation criteria used for feature detections.}
\tablabel{criteria}
\centering
\begin{tabular}{| l | p{0.65\linewidth} |}
\hline
Feature & Criteria for expected observability\\
\hline
Field boundary & The field should be visible in the image.\\
Ball & At least one third of the ball should be visible.\\
& The ball should be completely within the field boundary at $\leq\,$\SI{4}{\metre} from the robot.\\
Field lines & All points of the field line segment should be at most \SI{4.5}{\metre} from the robot.\\
& The line segment should be at least \SI{1}{m} in length.\\
Centre circle & At least one third of the circle should be visible.\\
Goal posts & The goal post should be $\leq\,$\SI{5}{m} from the robot.\\
& The base of the goal post should be visible.\\
& The goal post should have a minimum margin of \SI{30}{px} from the edges of the image.\\
Other robots & The feet of the robot should be completely black and visible.\\
\hline
\end{tabular}
\end{table}
\begin{table}[!tb]
\renewcommand{\arraystretch}{1.1}
\caption{Detection results for each type of feature.}
\tablabel{results}
\centering
\begin{tabular}{| l | c c c |}
\hline
Feature & Success rate & False positives & Frames\\
\hline
Field boundary & 93\% & 0 & 1482\\
Ball & 81\% & 11 & 457\\
Field lines & 57\% & 17 & 187\\ 
Centre circle & 63\% & 7 & 138\\ 
Goal posts & 48\% & 3 & 140\\
Other robots & 71\% & 0 & 239\\
\hline
\end{tabular}
\end{table}
\addtolength{\textheight}{-56mm}

\section{Conclusion and Future Work}
\seclabel{conclusion}

In this paper, we presented a monocular vision system for humanoid soccer robots 
that addresses the challenges posed by the new rules of the RoboCup 
Humanoid League. In order to tackle the challenging task of finding objects in a 
low color information environment, we proposed a learning approach for ball 
detection. For detecting goals, field lines, the center circle, and other robots 
we, proposed effective algorithms with low computational cost. 
Our methods have been evaluated on data from a RoboCup competition and in lab experiments.
The results indicate good detection performance and robustness to changes in lighting conditions.

In the future, we want 
to increase the maximum distance of ball detection, and utilize stereo vision 
for better distance estimation. Porting some of the system to run on a GPU is 
additional future work, along with developing a learning approach for more 
general robot detection, as the feet of other robots may not always be visible 
in the image.

\bibliographystyle{IEEEtran}
\bibliography{IEEEabrv,ms}

\end{document}